\documentclass[11pt]{article}
\usepackage[preprint]{acl}
\usepackage{times}
\usepackage{silence}
\usepackage{latexsym}
\author{
 \textbf{Lukas Hinterleitner\textsuperscript{1}},
 \textbf{Loris Schoenegger\textsuperscript{1,2}},
 \textbf{Benjamin Roth\textsuperscript{1,3}}
\\
\\
 \textsuperscript{1}Faculty of Computer Science, University of Vienna, Vienna, Austria\\
 \textsuperscript{2}UniVie Doctoral School Computer Science, University of Vienna, Vienna, Austria\\
  \textsuperscript{3}Faculty of Philological and Cultural Studies, University of Vienna, Vienna, Austria\\
\\
 \small{
   \textbf{Correspondence:} \href{mailto:contact@lukas-hinterleitner.at}{contact@lukas-hinterleitner.at}
 }
}

\usepackage[T1]{fontenc}
\usepackage{inputenc}
\usepackage{microtype}
\usepackage{inconsolata}
\usepackage{graphicx}
\usepackage{enumitem}
\usepackage{placeins}
\usepackage{amsmath}
\usepackage{amssymb}
\usepackage{booktabs}
\usepackage{multirow}
\usepackage{algpseudocode}
\usepackage{colonequals}
\usepackage{pifont}
\usepackage{subcaption}
\usepackage{mathtools}
\usepackage{algorithm2e}
\newif\ifhighlight
\highlightfalse     

\ifhighlight
  \usepackage{luacolor}
  \usepackage{lua-ul}
\fi

\newcommand{\revision}[2]{%
  \ifhighlight
    \highLight[#1!70!white]{#2}%
  \else
    #2%
  \fi
}
\title{Select or Project? Evaluating Lower-dimensional \\Vectors for LLM Training Data Explanations}

\begin{document}
\maketitle
\begin{abstract}
Gradient-based methods for instance-based explanation for large language models (LLMs) are hindered by the immense dimensionality of model gradients. In practice, influence estimation is restricted to a subset of model parameters to make computation tractable, but this subset is often chosen ad hoc and rarely justified by systematic evaluation.
This paper investigates if it is better to create low-dimensional representations by \textit{selecting} a small, architecturally informed subset of model components or by \textit{projecting} the full gradients into a lower-dimensional space. 
Using a novel benchmark, we show that a greedily selected subset of components captures the information about training data influence needed for a retrieval task more effectively than either the full gradient or random projection.
We further find that this approach is more computationally efficient than random projection, demonstrating that targeted component selection is a practical strategy for making instance-based explanations of large models more computationally feasible.
\end{abstract}

\section{Introduction}
Different modalities for explaining language model (LM) behavior in terms of causal influences have been proposed recently, including feature-based \citep{DBLP:journals/natmi/ChenCLL23, DBLP:conf/acl/Enguehard23}, or mechanistic explanations \citep{DBLP:conf/nips/MengBAB22, DBLP:journals/corr/abs-2307-09458}. 
While these approaches explain language model behavior in terms of inputs or architectural components, recent work has enabled the generation of instance-based explanations that provide insights grounded in the model's training data \citep{DBLP:conf/nips/HaraNM19, DBLP:conf/nips/PruthiLKS20, DBLP:conf/emnlp/GuoRHBX21, DBLP:journals/corr/abs-2308-03296}.
In this paper, we specifically focus on methods that identify influential training examples through the analysis of model gradients \revision{yellow}{, a signal widely exploited by XAI methods} \revision{cyan}{\cite[e.g.,][]{DBLP:conf/nips/PruthiLKS20, DBLP:conf/icml/ParkGILM23,DBLP:conf/icml/KohL17}}. By retrieving a set of examples that influenced a given model output, these methods enable model debugging and interpretability tasks \citep[e.g.,][]{DBLP:conf/icml/KohL17,DBLP:conf/nips/PruthiLKS20,DBLP:conf/emnlp/GuoRHBX21}.
However, their application to modern large language models is challenging due to high computational costs: for a 1B-parameter model, a single gradient requires over 4 GB of memory.
To make analysis tractable, it must often be restricted to a subset of model parameters \cite[e.g, the word-embedding layer:][]{DBLP:conf/nips/YehTSLR22}.
In existing work, this choice was often made ad hoc and rarely justified by systematic evaluation.
In this paper, we address this issue by investigating whether it is better to \textit{select} a sparse, structurally informed subset of model components or to \textit{project} the entire high-dimensional gradient into a dense, lower-dimensional space.
Specifically, we systematically compare two strategies for creating gradient surrogates:
An architecture-aware greedy layer selection algorithm proposed in this work, that selects a set of layer components whose combined gradients are most informative. And \textit{random projection}, a widely used, architecture-agnostic technique that produces a low-dimensional representation of the full gradient \citep{johnson_lindenstrauss}.

We evaluate using a \textbf{novel benchmark for instance-based explanations}, designed to test how well a given selection supports a retrieval task in which the original training example must be identified among alternative prompt–completion pairs generated using various strategies.
Our \textbf{main contributions} are threefold: 
\begin{enumerate}[nosep, topsep=0pt, label=(\arabic*), left=0pt]
\item  We propose an efficient retrieval-based benchmark for instance-based explanations.
\item We conduct a principled comparison of architecture-aware selection versus geometry-preserving projection for attribution.
\item We demonstrate that our targeted selection strategy can result in more accurate attributions while also being more efficient. 
\end{enumerate}

We find that a greedily selected subset of model components captures the required information about training data influence more accurately than either the full gradient or random projection. We also find that this targeted selection is more computationally efficient than random projection.
Our central insight is that a carefully chosen subset can provide a clearer and more discriminative signal for certain tasks, making it a practical strategy for instance-based explanations of large models.\footnote{We make our code available at \url{https://doi.org/10.5281/zenodo.18346666}.}
\begin{figure*}[htb!]
    \centering
    \includegraphics[width=1\textwidth]{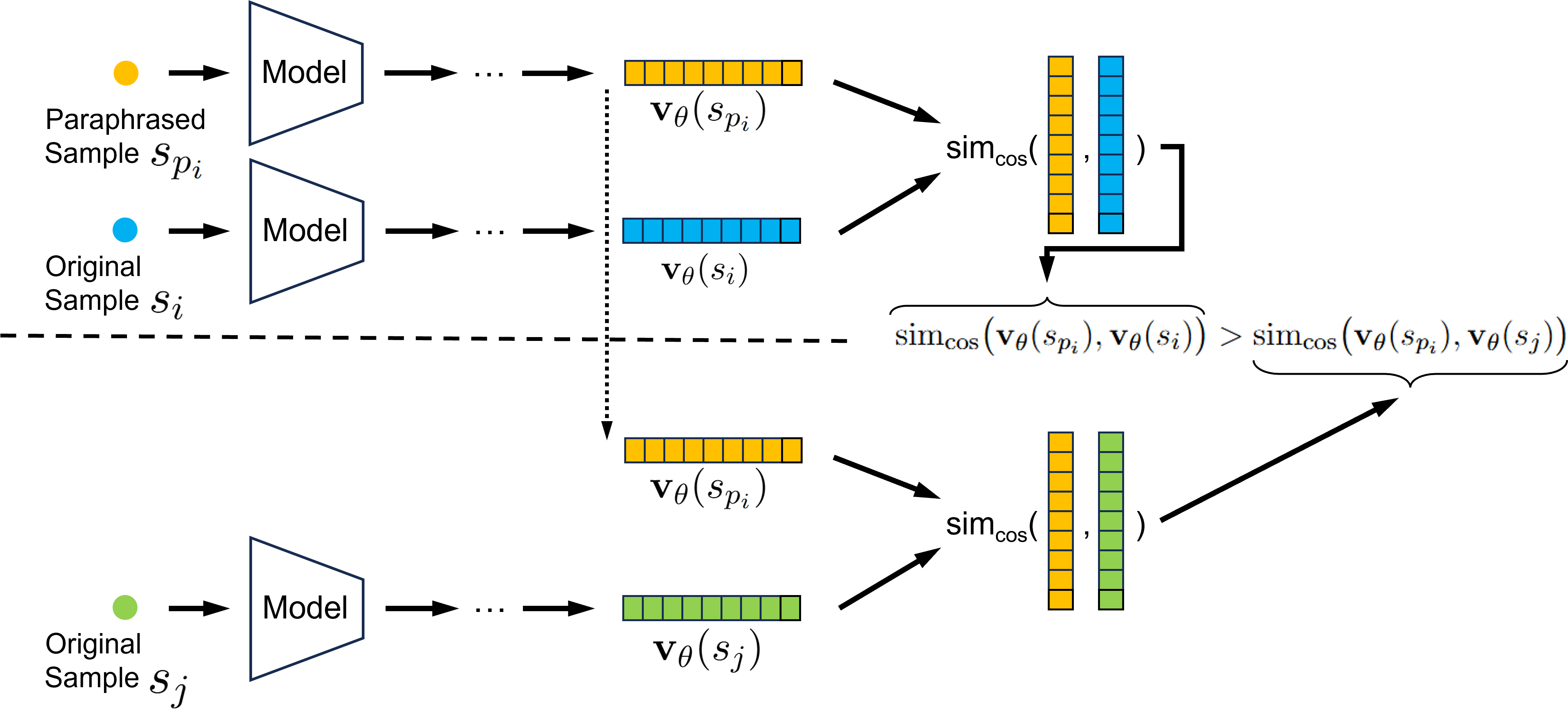}
    \caption{Overview of the evaluation setup for the \textit{paraphrased} setting. The gradient of a paraphrased sample $s_{p_i}$ is compared to gradients from its original counterpart $s_i$ and other candidates (e.g., $s_j$). The method succeeds at this retrieval task if the cosine similarity is highest for the original pair. The setup is identical for the \textit{generated} setting.}
    \label{fig:gradient_cosine_similarity}
\end{figure*}
\section{Related Work}
\paragraph{Instance-based explanation and attribution}
Instance-based explanation aims to identify influential training examples that explain a model's prediction. A dominant framework is \textit{leave one out influence} (the effect of removing a given training example has on the prediction), which can be approximated with re-training free methods such as \textit{influence functions} \citep{DBLP:conf/icml/KohL17,DBLP:journals/corr/abs-2308-03296}. Alternative approaches, such as TracInCP \citep{DBLP:conf/nips/PruthiLKS20}, aim to approximate training data influence \textit{during} the training process by comparing the test instance's loss gradient to training gradients at a set of model checkpoints.

\paragraph{The challenge of high-dimensional gradients}
These influence estimation methods involve comparisons of model gradients (and calculating inverse Hessian-vector products for influence functions), which remains computationally intractable for LLMs: recent applications therefore restrict computation to a subset of model parameters, such as the
multilayer perceptron (MLP) layers \citep{DBLP:journals/corr/abs-2308-03296} or LoRA layers \citep{kwon_datainf_2023,schoenegger2026compactexamplebasedexplanationslanguage}.
Another common choice is to restrict influence estimation to the model's final layers \citep{DBLP:conf/aaai/SchioppaZVS22, DBLP:conf/emnlp/AkyurekBLXTAG22, DBLP:journals/corr/abs-2308-03296}. However, \citet{DBLP:conf/nips/YehTSLR22} find this to be suboptimal for LLMs due to a \textit{cancellation effect} in deeper layers, and propose using early layer components instead, for example, the more stable word embedding gradients. 
An alternative line of work projects full model gradients to a lower-dimensional space where pairwise distances between examples are preserved \cite[e.g.,][]{DBLP:conf/icml/ParkGILM23,xia_less_2024,lin-etal-2024-token}.
Our work systematically compares the utility of \emph{all} architectural components and random projection to determine which strategy provides the most informative signal for retrieval.

\paragraph{Evaluation of training data explanations}
\citet{DBLP:journals/corr/abs-2308-03296} evaluate their approximation of influence functions by reporting agreement with the \textit{proximal Bergmann response function}, whose approximation does not require re-training. However, as they note themselves, it remains unclear whether this is a useful measure of data influence.
\citet{DBLP:conf/emnlp/AkyurekBLXTAG22} evaluate by introducing facts with the fine-tuning data, that the model of interest did not know before, and then verify whether influence estimation methods correctly attribute importance to this fine-tuning set when queried about them.
Such evaluation strategies require expensive additional training steps that are infeasible with LLMs. 
Our evaluation setup is most similar to \citet{DBLP:conf/iclr/WangMS025} who use paraphrased data to test the performance of influence estimation methods like \textit{Data Shapley} \citep{pmlr-v97-ghorbani19c} and influence functions by identifying the original sources of rewritten text. 
Similarly, our evaluation setup is re-training free; we use it to systematically compare dimensionality reduction and gradient selection strategies for influence estimation.
\section{Method: Gradient Similarity for Instance-based Explanations}
Instance-based explanation methods must be capable of distinguishing training examples that influence model output from those that do not. Popular methods do so based on model loss gradients.
As we will outline in Section \ref{eval_via_retirval} we frame this requirement as a retrieval problem:
To study the quality of different low-dimensional representations, our setup measures how well they enable one to identify a given test example among distractors.
We introduce a novel low-dimensional representation strategy in Section \ref{novel_selection}  and describe our setup for random projection in Section \ref{rp}. Our evaluation approach is (1) \textbf{local}, targeting decisions at the instance level (each training sample/instance processed individually); (2) \textbf{static}, using the final model checkpoint without reconstructing training dynamics; and (3) \textbf{model-invariant}, applicable to any model that provides per-sample gradients.

\subsection{Gradient Representation and Similarity}
Let $D = \{s_i\}_{i=1}^N$ be a subset of a fine-tuning dataset for a given LLM with parameters $\theta$. The training objective is to minimize a loss function, $\mathcal{L}(s_i, \theta)$. The gradient for a single sample $s_i$ with respect to the final, fine-tuned model parameters $\theta_T$, denoted $\nabla_{\theta} \mathcal{L}(s_i, \theta_T)$, indicates the direction in the parameter space that would reduce the loss for that specific sample. It serves as a representation of the sample's influence on the model before being utilized by an optimizer, like AdamW~\cite{DBLP:conf/iclr/LoshchilovH19}.
The model's parameters $\theta$ are a collection of component tensors $\{\mathbf{W}^{(l,k)}\}$, where $l$ indexes the layer and $k$ the specific component within it (e.g., query, key, or value projections in an attention block; MLP layers). Let $\mathcal{W}$ be the set of all such index pairs $(l,k)$. To create a single vector for comparison, we first calculate the gradient w.r.t each component, $\nabla_{\mathbf{W}^{(l,k)}}\mathcal L(s_i,\theta_T)$ and secondly, flatten each gradient before combining all of them into a single vector $\mathbf{v}_\theta(s_i) \in \mathbb{R}^{M}$ (we refer to this as the full model gradient or full gradient), where $M$ is the total number of parameters (see Appendix~\ref{app:gradient_flattening} for more details on gradient flattening). We use cosine similarity $\text{sim}_{\text{cos}}(\cdot,\cdot)$ to then compare gradient vectors (see Appendix~\ref{app:cosine_reconstruction}, on how cosine similarity suits our approach).
\subsection{Evaluation via Retrieval}\label{eval_via_retirval}
\revision{yellow}{A sample's loss gradient encodes how a training step on it would adapt the model's parameters. Gradient-based attribution methods build on this: TracIn-style methods \citep{DBLP:conf/nips/PruthiLKS20} score influence directly by the dot product between training- and test-sample loss gradients, the first-order effect of a training step on the test loss, while influence functions \citep{DBLP:conf/icml/KohL17} additionally weight this inner product by the inverse Hessian. A paraphrase should adapt the model similarly to its original, so the original should be identified as most influential; our retrieval task turns this into a testable necessary condition: a representation that cannot rank the original above closely related alternatives cannot yield reliable influence rankings.}

To study the utility of different representations, such as using the \textit{full} gradient for influence estimation (Section 3.2), \textit{selecting} a subset of model parameters (Section 3.3), or random \textit{projection} (Section 3.4), we consider the following retrieval task:
We first derive two different query datasets ($D_p$ and $D_m)$ from the original fine-tuning dataset $D$ by paraphrasing each individual training sample. 
We use two different strategies: 
\textbf{Paraphrased ($D_p$)}, where each sample $s_{p_i} \in D_p$ is a paraphrase of $s_i \in D$, preserving semantics while altering lexical form.
And \textbf{Model-Generated ($D_m$)}, where for each $s_i$, we paraphrase its prompt and use the fine-tuned model itself to generate a new completion, creating $s_{m_i} \in D_m$. This tests the model's understanding of its own outputs for semantically similar inputs.
For simplicity, we describe our method using the paraphrased set $D_p$ throughout the paper, the process for $D_m$ is identical. 

Our initial assumption is that the cosine similarity between the gradients of a paraphrased sample $s_{p_i}$ and its original counterpart $s_i$ is higher than the similarity between the gradients of $s_{p_i}$ and an other sample from the training data $s_j \in D$. Specifically, we test whether: 
\begin{equation}
    \text{sim}_{\text{cos}}(\mathbf{v}_\theta(s_{p_i}),\mathbf{v}_\theta(s_i)) > \text{sim}_{\text{cos}}(\mathbf{v}_\theta(s_{p_i}),\mathbf{v}_\theta(s_j)), \notag
\end{equation}
where $i \neq j$. If this assumption holds, then we should also be able to retrieve the original counterpart for every {paraphrased} sample $s_{p_i}$ (Figure~\ref{fig:gradient_cosine_similarity}).
For each $s_{p_i} \in D_p$ , we therefore test if its original $s_i$ is ranked highest among a set of candidates $\mathcal{C}_i(b) \subset D$. This set is formed by retrieving the top-$b$ most lexically similar samples to $s_{p_i}$ from $D$ using BM25 \citep{DBLP:journals/ftir/RobertsonZ09}.\footnote{\revision{yellow}{Given the user prompts have a short median length of 17 words, simpler lexical-overlap measures than BM25 would likely suffice; the original counterpart is always included in the candidate set (Appendix~\ref{app:candidate_creation}).} }This candidate set is created to reduce the search space and hence computing fewer gradients.
Our evaluation metric is retrieval accuracy:
\begin{equation*}
\resizebox{0.9\linewidth}{!}{$
\begin{aligned}
    \operatorname{acc}^{(b)}_{\theta}(D_p, D) &=\frac{1}{N}\sum_{i=1}^{N} \mathbb{I}\bigg[  \text{sim}_{\text{cos}}(\mathbf{v}_{\theta}(s_{p_i}), \mathbf{v}_{\theta}(s_i)) \\
    &> \max_{j \in \mathcal{C}_i(b) \setminus \{i\}} \text{sim}_{\text{cos}}(\mathbf{v}_{\theta}(s_{p_i}), \mathbf{v}_{\theta}(s_j)) \bigg]
\end{aligned}
$}
\end{equation*}
where $\mathbb{I}[\cdot]$ is the Iverson bracket.

\subsection{Greedy Component Selection (\textit{Select})}\label{novel_selection}
Our approach constructs a smaller, architecturally-informed surrogate gradient by greedily selecting a subset of model components up to a given budget that are most informative for the retrieval task by leveraging the linearity of the dot product: The dot product of concatenated vectors is simply the sum of the dot products of the constituent vectors. This allows us to reconstruct the cosine similarity for any subset of component indices $\mathcal{S} \subseteq \mathcal{W}$ by summing pre-computed scalar values. Let $\mathbf{v}_{\mathcal{S}}(s_i)$ be the gradient vector formed by concatenating component gradients for indices in $\mathcal{S}$ (see Appendix~\ref{app:cosine_reconstruction}, for a more detailed view on the cosine similarity reconstruction). We employ a greedy forward algorithm that iteratively adds the component yielding the greatest improvement in retrieval accuracy. Let $\mathcal{S}$ be the set of already selected layer component indices. At each step, we choose the next component index $(l^*, k^*)$ to add:
\begin{equation*}
(l^*,k^*) = \arg\max_{(l,k) \in \mathcal{W} \setminus \mathcal{S}} \operatorname{acc}^{(b)}_{\mathcal{S} \cup \{(l,k)\}}(D_p, D)
\end{equation*}
This process is efficient as it operates entirely on pre-computed dot products, avoiding re-computation or storage of high-dimensional gradients during search.

\subsection{Random Projection (\textit{Project})}\label{rp}
Following previous work \cite[e.g.,][]{DBLP:conf/icml/ParkGILM23,xia_less_2024,lin-etal-2024-token}, we create a dense surrogate gradient by projecting the full gradient into a lower-dimensional space using an efficient GPU-optimized implementation from \citet{DBLP:conf/icml/ParkGILM23}. A random projection matrix $\mathbf{R} \in \mathbb{R}^{d \times M}$ (where $d \ll M$) maps the full gradient $\mathbf{v}_\theta(s_i)$ to a smaller vector $\mathbf{p}(s_i) \in \mathbb{R}^d$:
\begin{equation*}
\mathbf{p}(s_i) \coloneqq \frac{1}{\sqrt{d}} \mathbf{R} \, \mathbf{v}_\theta(s_i).
\end{equation*}
Due to the immense size of $\mathbf{R}$ and $\mathbf{v}_\theta(s_i)$, we cannot materialize them directly. Instead, we perform the projection in a component-wise fashion. For each component $\mathbf{W}^{(l,k)}$, we use a separate random matrix $\mathbf{R}^{(l,k)}$, where the projection dimension for that component is proportional to its parameter count. The final low-dimensional representation is the concatenation of these individual projections. This provides an architecture-agnostic baseline that aims to preserve the geometric structure of the full gradient per sample $\mathbf{v}_\theta(s_i)$.
\section{Experimental Setup}
\label{sec:experimental_setup}

\subsection{Model and Data}
We conduct our experiments using \texttt{AMD-OLMo-1B-SFT} \citep[][Apache license 2.0]{DBLP:conf/acl/GroeneveldBWBKT24, AMD-OLMo}, a 1.2B parameter decoder-only transformer model. We choose this model as it's training data is made available in full. Specifically, it was pre-trained on a 1.3T token subset of the \texttt{Dolma} dataset \citep{DBLP:conf/acl/SoldainiKBSAABC24}, and fine-tuned on the \texttt{Tülu 2} \citep{DBLP:journals/corr/abs-2311-10702}, \texttt{OpenHermes-2.5} \citep{oh}, 
\texttt{WebInstructSub} \citep{yue2024mammoth2}, and \texttt{Code-Feedback} \cite{zheng-etal-2024-opencodeinterpreter} datasets. 

Including the input embedding layer, we consider a total of $|\mathcal{W}| = 113$ distinct component tensors for our analysis. This comprises 16 layers, each containing seven trainable weight matrices: four for the attention mechanism (Q, K, V, and O projections) and three for the MLP block (Gate, Up, and Down projections).

Our base dataset for the evaluation setup, $D$, is the \texttt{LIMA} \citep[][CC BY-NC-SA license]{DBLP:conf/nips/ZhouLX0SMMEYYZG23} subset of \texttt{Tülu 2} \citep{DBLP:journals/corr/abs-2311-10702}\revision{yellow}{: we use all samples ($\approx$1k) except for a small fraction of multi-turn dialogues, yielding $N=988$ single-turn prompt-generation pairs. We do not apply any further sampling or filtering. Prompts are typically short (median 21, mean 54 tokens); generations are longer (median 369, mean 587 tokens)}. All samples are formatted using the model's chat template, which wraps user instructions in \texttt{<|user|>} and marks the generation point with \texttt{<|assistant|>}, ensuring the input aligns with the model's fine-tuning format. We provide $s_0$ as an example below:
\begin{quote}
\footnotesize
\begin{verbatim}
<|user|>
Can brain cells move? ...
<|assistant|>
The question is relatively broad...
\end{verbatim}
\end{quote}

\subsection{Implementation Details}
The sheer scale of a billion-parameter model introduces significant computational hurdles. The full flattened gradient for a single sample, $\mathbf{v}_\theta(s_i)$, is a vector in $\mathbb{R}^{1.2B}$, requiring approximately 4.7 GB of storage, which makes naively storing all gradients infeasible.
To make the retrieval task tractable, the candidate set $\mathcal{C}_i(b)$ is limited to $b=5$ samples. This targeted approach, pre-filtered with BM25, reduces the number of required gradient similarity computations by over 99\%. 
\revision{cyan}{}
\paragraph{Intermediate dot product storage}
To implement the \textit{Select} method without storing high-dimensional vectors, we compute and store only their intermediate, component-wise dot products. For each component $(l,k) \in \mathcal{W}$ and every candidate pair of samples $(s_i, s_j)$, we pre-compute the scalar values $\mathbf{v}_{\mathbf{W}^{(l,k)}}(s_i)^{\top}\mathbf{v}_{\mathbf{W}^{(l,k)}}(s_j)$ and the self-dot-products for both $s_i$ and $s_j$. These values can be aggregated post-hoc to reconstruct similarity scores for any subset of components (see Appendix~\ref{app:cosine_reconstruction} for additional information).
\paragraph{Layer-wise random projection}
The \textit{project} baseline is implemented using a component-wise approach because a naive projection of the entire 1.2B-parameter gradient is impossible, as the projection matrix itself would require petabytes of storage. We instead apply a separate random projection to each of the 113 vectorized component gradients. This was implemented using the \texttt{CudaProjector} from the \textit{TRAK} library~\citep{DBLP:conf/icml/ParkGILM23}.

\paragraph{Evaluation dataset construction}
The query datasets were constructed using specific models. For the paraphrased set ($D_p$), we used GPT-4o-mini to paraphrase every sample in the base dataset $D$. For example, $s_{p_0}$ generated from $s_0$ is:
\begin{quote}
\footnotesize
\begin{verbatim}
<|user|>
Are brain cells capable of moving? ...
<|assistant|>
The inquiry is quite extensive...
\end{verbatim}
\end{quote}
For the model-generated set ($D_m$), we fed the paraphrased prompts from $D_p$ to the fine-tuned OLMo model to create corresponding generations.
\section{Results and Analysis}\label{sec:results}
We evaluate in two settings: \textit{paraphrased}, where inputs are semantically similar, and \textit{model-generated}, where the model produces a novel generation to the paraphrased prompt.

\subsection{Full Gradient Performance}
First, we evaluate the full gradient as a baseline. As shown in Table \ref{tab:single_component_accuracy}, the full gradient is highly effective in the paraphrased setting ($D_p$), achieving near-perfect retrieval accuracy of $0.993$. However, its performance collapses to $0.218$ on the more challenging model-generated dataset $D_m$, which is only marginally better than the random baseline of $0.20 = 1/5$, when considering a candidate set with size $b=5$.
A qualitative analysis of the few failure cases in the paraphrased setting reveals that for small amount ($<1\%$) of instruction-based prompts, the paraphrasing LLM \textit{executed} instructions rather than rephrasing them. For example:
\begin{quote}
\footnotesize
\textbf{Original Prompt:} \texttt{translate into English: "Der Zug..."} \\
\textbf{Paraphrased Prompt:} \texttt{The train arrives in Frankfurt...}
\end{quote}
The paraphrased prompt contains the \textit{result} of the instruction, not the instruction itself. This fundamentally alters the task, shifting the token distribution and the resulting gradient, which likely causes the retrieval to fail.

\subsection{Single Layer Component Performance}
Interestingly, gradients from a small subset of model components perform as well as, or even better than, the full model gradient. Table \ref{tab:single_component_accuracy} shows the mean retrieval accuracy for each component type, averaged across all layers:
In the paraphrased setting, most components achieve near-perfect accuracy, indicating that the signal is broadly spread. In the model-generated setting, however, performance is concentrated in specific components. The MLP \textit{gate} and \textit{up} projections carry the most informative signal ($\approx 0.26$), while the attention's \textit{key} ($0.208$) and \textit{value} ($0.198$) projections contribute little useful information for retrieval, with the latter performing worse than random chance.

\begin{table}[ht!]
\centering
\footnotesize
\begin{tabular}{@{}lcc@{}}
\toprule
\textbf{Component Type} & \textbf{Paraphrased} & \textbf{Model-Generated} \\
& (Mean Acc.) & (Mean Acc.) \\
\midrule
\textit{MLP Components} & & \\
MLP Down-Proj & \textbf{0.993} & 0.235 \\
MLP Gate-Proj & 0.992 & \textbf{0.256} \\
MLP Up-Proj & 0.991 & 0.247 \\
\midrule
\textit{Attention Components} & & \\
Attn Output-Proj & 0.992 & 0.245 \\
Attn Query-Proj & 0.976 & 0.255 \\
Attn Value-Proj & 0.98 & 0.198 \\
Attn Key-Proj & 0.916 & 0.208 \\
\midrule
\textit{Embedding} & & \\
Embedding Layer & \textbf{0.993} & 0.231 \\
\midrule
\textbf{Full Model Grad.} & \textbf{0.993} & 0.218 \\
\bottomrule
\end{tabular}
\caption{Mean retrieval accuracy per component type. In the model-generated setting, MLP components are most informative, while some attention components fail.}
\label{tab:single_component_accuracy}
\end{table}

\begin{figure}[ht!]
    \centering
    \includegraphics[width=1\columnwidth,trim={0.7cm 0 1.1cm 0}, clip]{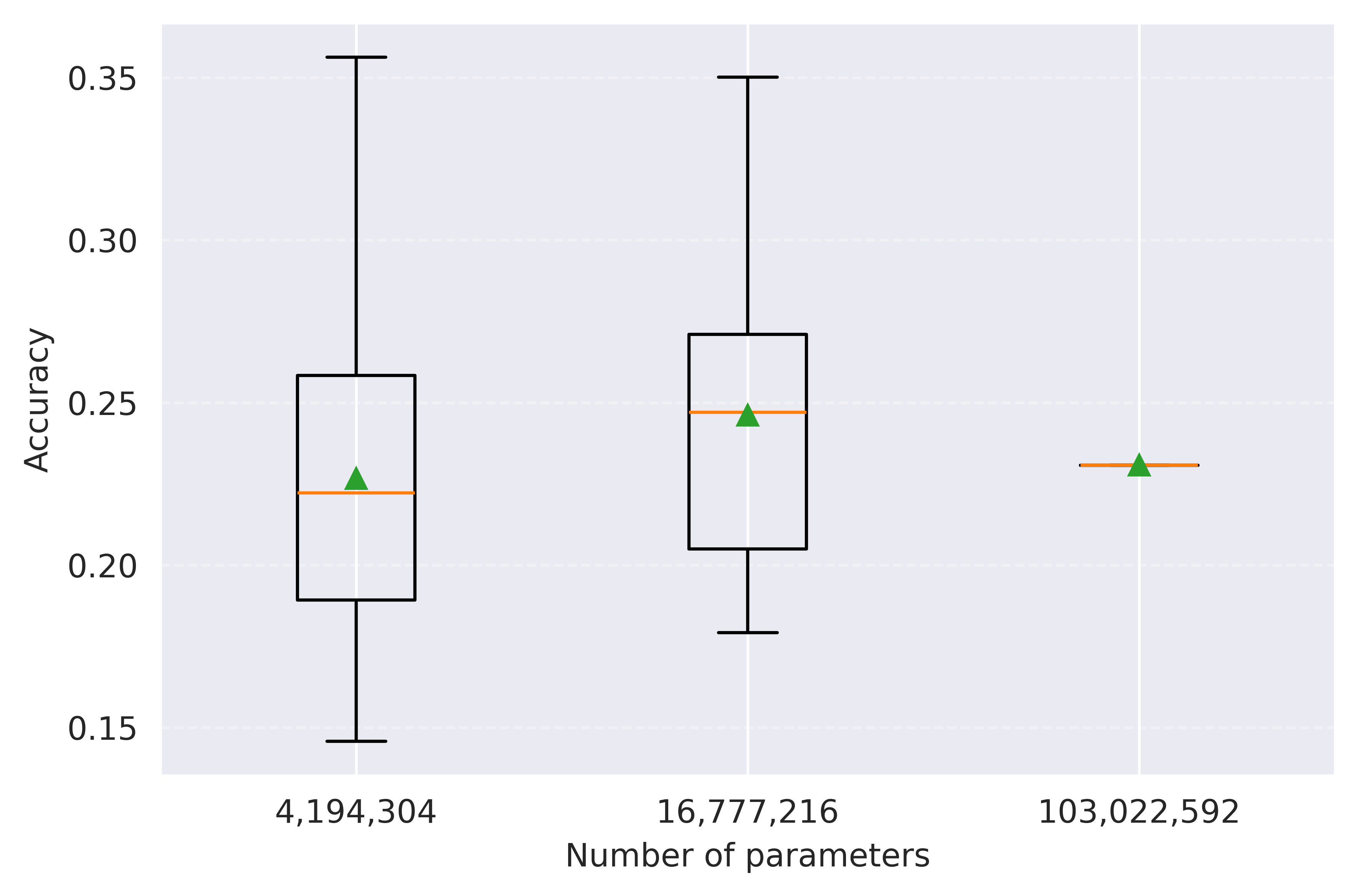}
    \caption{Component accuracy vs. parameter count. Each boxplot groups all components of a specific type, which share the same size. \textbf{Left}: Attention projections. 
    \textbf{Center}: MLP components. 
    \textbf{Right}: Embedding layer.
    }
    \label{fig:accuracy_boxplot}
\end{figure}

\paragraph{The role of component size and function}
We investigate if parameter count correlates with performance. As shown in Figure \ref{fig:accuracy_boxplot}, there is no clear monotonic relationship, as components of the same type share the same parameter count. The three boxplots correspond to the attention projections (left, $4.2$M parameters), the MLP layers (center, $16.8$M parameters), and the embedding layer (right, $103$M parameters). The medium-sized MLP components on average outperform both the much larger embedding layer and the smaller attention projections. Furthermore, within the attention mechanism, all projection components have the same size, yet their individual performance varies significantly. This suggests that a component's functional role is a more critical determinant of its explanatory power than its parameter count.

\begin{figure*}[ht!]
    \centering
    \includegraphics[width=1\textwidth]{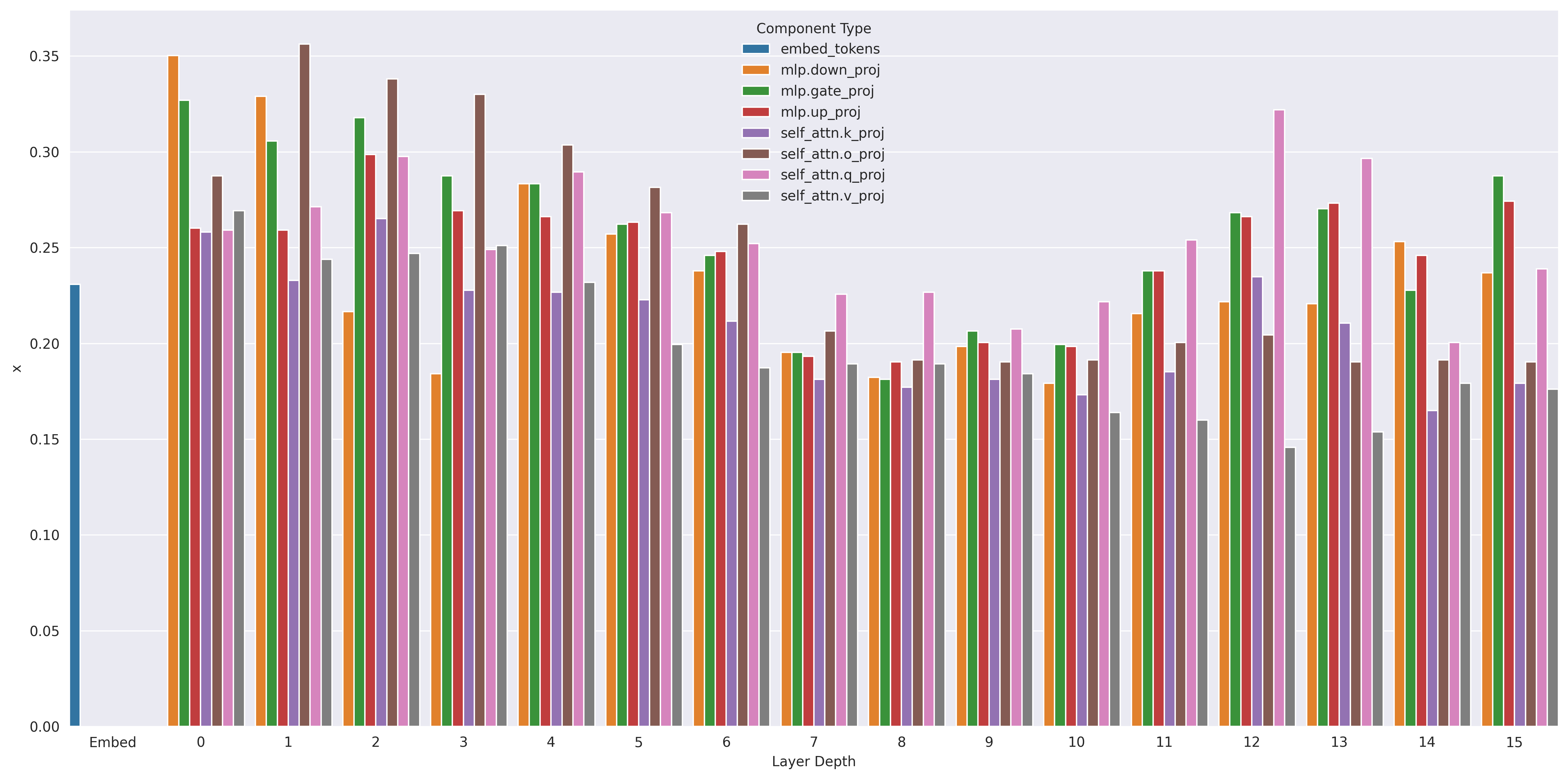}
    \caption{Accuracy vs. layer depth for the model-generated setting. Each color represents a layer component.}
    \label{fig:accuracy_depth}
\end{figure*}

\paragraph{Performance by model depth}
The accuracy distribution of influential gradient components across model depth differs substantially between settings. While accuracy is high and stable across all layers in the paraphrased setting, the model-generated one reveals a distinct pattern as evident in Figure~\ref{fig:accuracy_depth}. Performance is strongest in the initial and final layers, with a noticeable dip in the middle layers. This suggests that while early layers (near the input representation) and late layers (near the output logits) retain some alignment with the original sample, intermediate layers, which perform more abstract semantic transformations \citep{DBLP:conf/acl/TenneyDP19, DBLP:conf/acl/JawaharSS19}, appear less informative when the model generates a novel response. MLP layer components also regain performance in later layers, while the attention components (except Q) remain equal or worse. This is in line with the observations of \citet{DBLP:conf/nips/YehTSLR22}, who also observe that early layers appear more informative than later ones for training data attribution.

\subsection{Accuracy vs. Full Gradient Alignment}
Given that single gradient components can be highly informative, we explore whether a small \textit{subset} of components can outperform the full gradient in retrieval accuracy. Furthermore, we assess how well a \textit{subset} of components aligns with the full gradient by measuring how similar their cosine similarities are to those of the full gradient.

\begin{figure*}[ht!]
    \centering
    \begin{subfigure}{0.49\textwidth}
        \includegraphics[width=\textwidth]{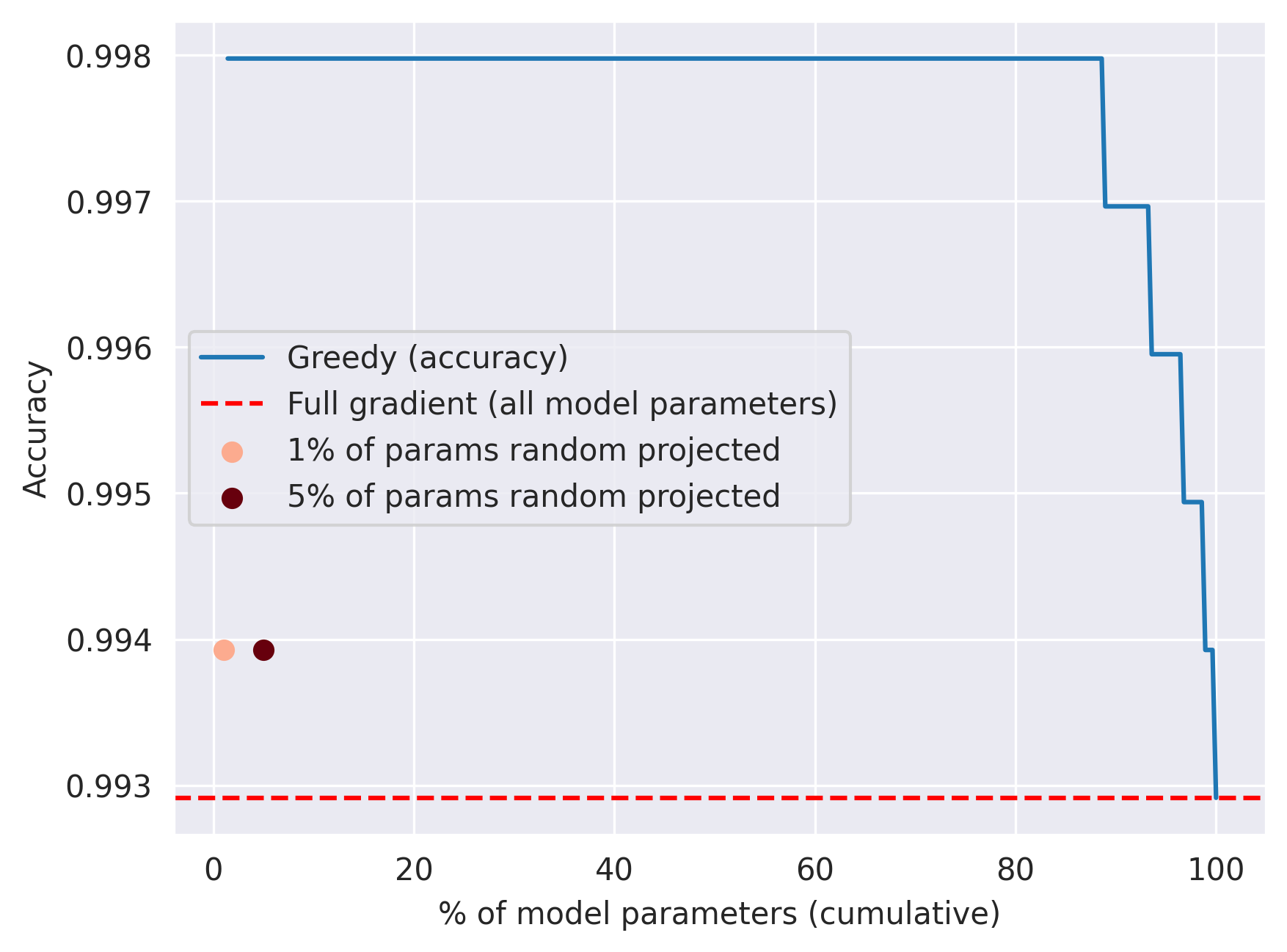}
        \caption{Retrieval Accuracy (Paraphrased)}
        \label{fig:accuracy_paraphrased}
    \end{subfigure}
    \hfill
    \begin{subfigure}{0.49\textwidth}
        \includegraphics[width=\textwidth]{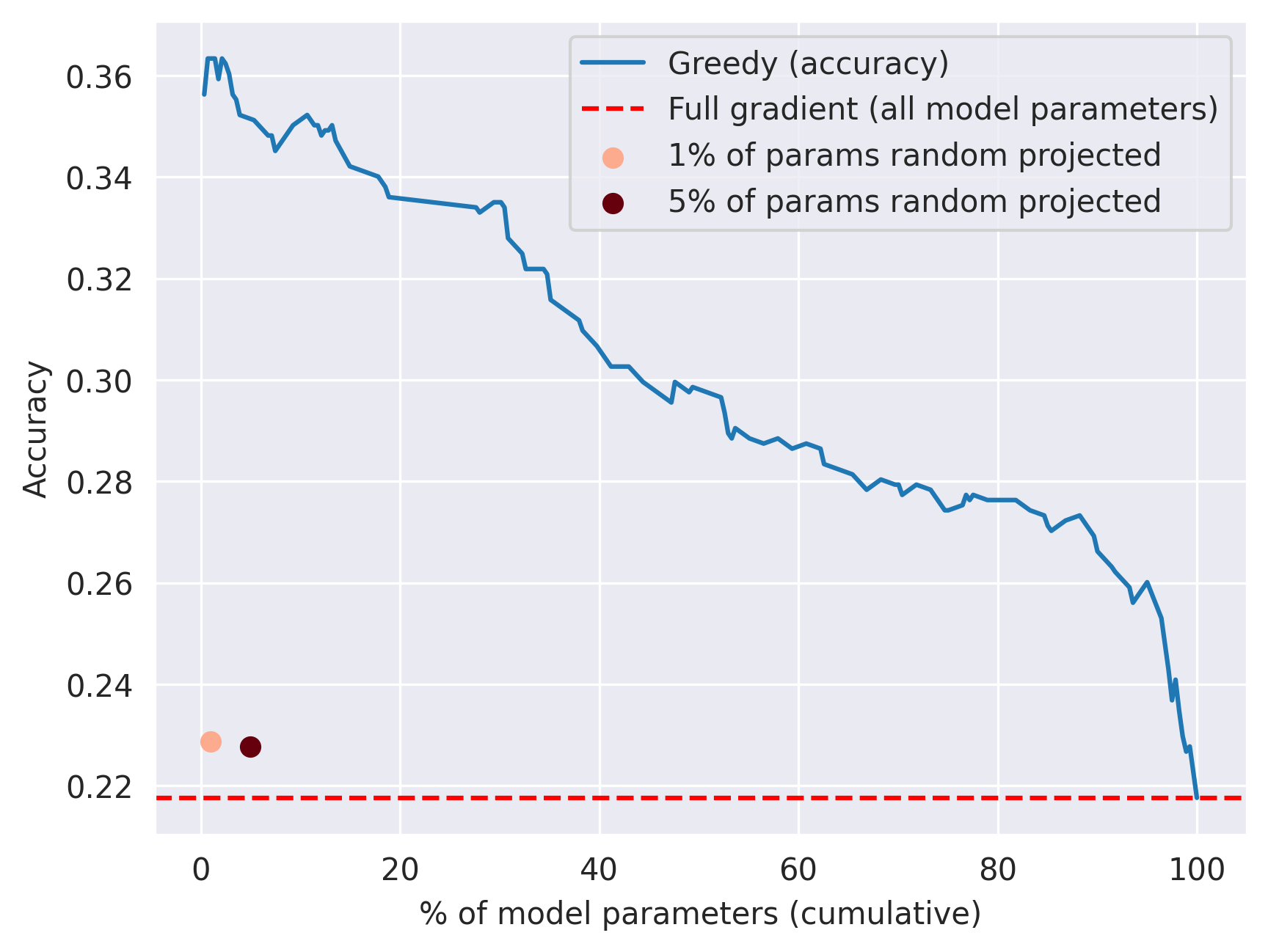}
        \caption{Retrieval Accuracy (Model-generated)}
        \label{fig:accuracy_model_generated}
    \end{subfigure}
    \caption{Greedy component selection (optimized for accuracy) vs. Random Projection. The x-axis shows the cumulative fraction of parameters utilized. In both settings, a small, greedily-selected subset outperforms the full gradient (dashed line) and random projection.}
    \label{fig:greedy_vs_rp}
\end{figure*}

\paragraph{Selection by retrieval accuracy}
Figure~\ref{fig:greedy_vs_rp} shows that greedily selecting components based on retrieval accuracy identifies a small subset that significantly outperforms both random projection and the full gradient. In the paraphrased setting, accuracy peaks at $0.998$ with less than $5\%$ of parameters for the greedy selection. This demonstrates a key finding: certain components of the full gradient appear noisy for the task at hand, and retrieval can be improved by filtering out less-informative elements. Even in the challenging model-generated setting, a small subset of components achieves an accuracy of $\approx 0.36$, nearly double that of the full gradient.

\begin{figure*}[ht!]
    \centering
    \begin{subfigure}[b]{0.49\textwidth}
        \centering
        \includegraphics[width=\textwidth]{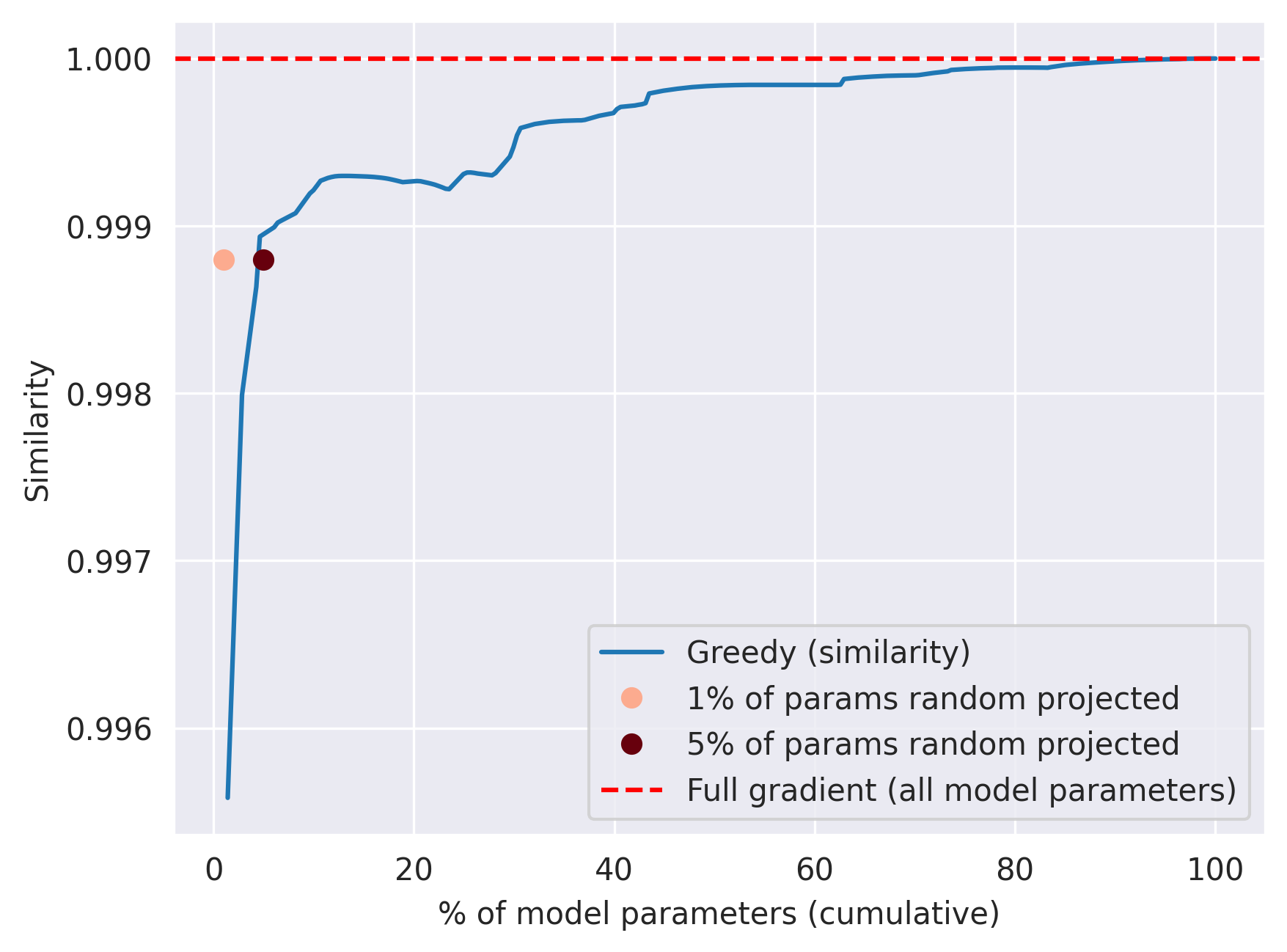}
        \caption{Similarity (Paraphrased)}
        \label{fig:greedy_layer_selection_paraphrased}
    \end{subfigure}
    \hfill
    \begin{subfigure}[b]{0.49\textwidth}
        \centering
        \includegraphics[width=\textwidth]{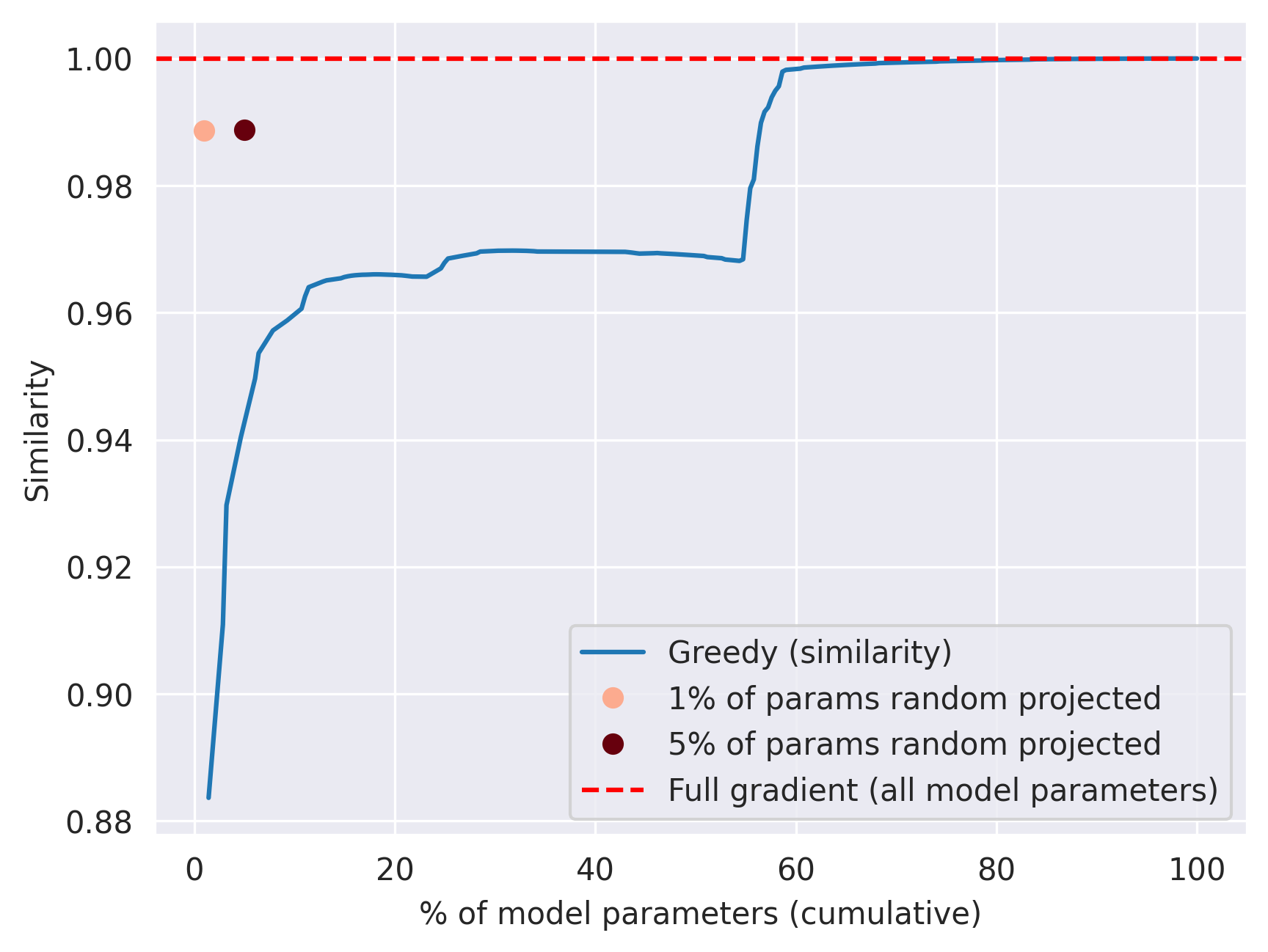}
        \caption{Similarity (Model-generated)}
        \label{fig:greedy_layer_selection_model_generated}
    \end{subfigure}
    \caption{Greedy component selection vs. random projection, optimizing for similarity to the full model gradient. While random projection quickly replicates the full gradient's geometry, this does not yield the best retrieval accuracy (cf. Figure \ref{fig:greedy_vs_rp}).}
    \label{fig:greedy_layer_selection}
\end{figure*}

\paragraph{Selection by similarity to the full gradient}
In contrast, when we select components to maximize cosine similarity with the full gradient scores (Figure~\ref{fig:greedy_layer_selection}), we observe that random projection is highly effective, achieving $\gtrsim 0.999$ similarity with just $1\%$ of parameters in the paraphrased setting. Also in the model-generated case, random projections outperform component selection. However, this high fidelity does not translate to optimal retrieval performance. This confirms that preserving the full gradient's global geometry is a different, and for our task less effective, objective than identifying the components most salient for retrieval.

\subsection{Analysis of Computational Cost}
The computational requirements of our proposed architecture-aware selection strategy is substantially lower than performing random projection: precomputing dot products for all components took approximately $4$ hours. The subsequent greedy selection process finished in a few minutes. In contrast, random projection baselines (without caching) were computationally more intensive, requiring over $900$ hours. However, due to storage limitations, gradient-caching was infeasible. For completeness, we calculated how much faster random projection would be if gradient-caching was applied without adding any read-write costs (see results in Table~\ref{tab:execution_times}). All computations were performed on NVIDIA H100 GPUs.
\begin{table}[htb!]
\centering
\footnotesize
\begin{tabular}{@{}lc@{}}
\toprule
\textbf{Computation Method} & \textbf{Time (Hours)} \\
\midrule
Dot-Product Pre-computation & $\sim$4 \\
Greedy Selection (Post-processing) & $\sim$0.03 \\
\textbf{Total for Greedy Method} & \textbf{$\sim$4.03} \\
\midrule
1\% RP Baseline (without caching) & $\sim$ 158 \\
5\% RP Baseline (without caching) & $\sim$ 765 \\
\textbf{Total RP Baselines (without caching)} & $\gtrsim$ 900 \\
\midrule
1\% RP Baseline (with caching) & $\sim$ 53 \\
5\% RP Baseline (with caching) & $\sim$ 255 \\
\textbf{Total RP Baselines (with caching)} & $\gtrsim$ 300 \\
\bottomrule
\end{tabular}
\caption{Computational cost comparison for the paraphrased setting. The greedy selection method remains substantially more efficient than random projection (RP), even if ignoring storage costs of gradient-caching.}
\label{tab:execution_times}
\end{table}
\section{Discussion}
Our findings challenge the notion that the full model gradient is the best source of information for instance-based explanation and reveal how well different layer components encode instance-specific information in our retrieval task. We provide a set of insights to improve future explanation setups.

\paragraph{The suitability of the full gradient}
The full gradient is often outperformed by some individual component gradients in the \textit{paraphrased} setting. Smaller subsets not only perform better in terms of accuracy but are also smaller, and therefore easier to store and retrieve. In the \textit{model-generated} setting, the full gradient is a weak baseline, with accuracy collapsing to 0.218 (vs. 0.2 random chance). Selecting a small subset of parameters consistently outperforms the full gradient, proving that its signal is diluted by weaker layer components in the case of instance-based methods.

\paragraph{Insights from analyzing gradient components}
Layer-wise analysis reveals where instance-specific information is encoded:
We observe that not all layers contribute equally. Specifically, MLP blocks and attention Q/O projections carry more information than K/V projections, regardless of size for the task we consider. This highlights the significance of a component's \textit{functional role} over its parameter count. We furthermore observe that greedy selection shows a non-monotonic link between parameter budget and accuracy. A small and optimal subset of components achieves competitive performance, while adding more components appears to introduce noise and degrades accuracy toward the full-gradient baseline. Lastly, influence distribution across depth varies by setting. In the model-generated setting, early and late layers appear most useful on average, while middle layers underperform. However, in later layers only MLP and attention Q blocks perform well.

\subsection{To Select or to Project?}
Our evaluation indicates that not all components are equally beneficial when retrieving relevant training samples, suggesting that subset selection should be preferred over random projection, which does not account for the informativeness of components during retrieval. Restricting influence estimation to a subset of layers can retain retrieval accuracy while improving computational efficiency. Consistent with this, our simple greedy parameter selection approach outperforms common random projection strategies.
Lastly, with appropriate adaptation to task specifics, this approach could also support applications that rely on similar loss-gradient comparisons, such as data cleaning- or pruning methods.
\section{Conclusion}
This work systematically evaluated strategies for reducing the computational requirements of gradient-based methods for instance-based explanations, comparing the targeted selection of a small subset of model components to random projection, and the use of the full gradient. We find that our proposed method of selecting a compact, structurally informed subset consistently outperforms full gradients and projections in both accuracy and efficiency.

Our results highlight that a small, carefully chosen set of components can be sufficient to capture the instance- and task-specific information required for generating instance-based explanations. Consequently, our work demonstrates that architecture-aware selection is an effective strategy to making instance-based explanations for large models more computationally feasible.
Equipped with our evaluation setup, future work should further explore efficient strategies for component selection.

\section*{Limitations}
Our experiments are conducted with a 1.2B-parameter model and its fine-tuning dataset, as obtaining random projections for the full gradient already requires over 300 GPU hours in our setup. Given sufficient resources, one could explore generalization across several axes, for example, analyzing different model architectures, scales, and data domains. 
Additionally, future work should extend analysis to \textit{dynamic} attribution methods that track influence throughout the training process at multiple checkpoints \cite{hammoudeh_training_2024}.

\section*{Acknowledgments}
This research has been funded by the Vienna Science and Technology Fund (WWTF) [10.47379/VRG19008] "Knowledge-infused Deep Learning for Natural Language Processing".
We also thank \citet{lin-etal-2024-token} for providing the illustration templates of their paper \textit{Token-wise Influential Training Data Retrieval for Large Language Models}, on which Figure~\ref{fig:gradient_cosine_similarity} builds.

\bibliography{custom}

\appendix

\FloatBarrier
\section{Methodological Details}\label{app:method_details}

\subsection{Gradient Flattening}\label{app:gradient_flattening}
To create a single vector for comparison, we first compute the gradient w.r.t. each component tensor, $\nabla_{\mathbf{W}^{(l,k)}}\mathcal L(s_i,\theta_T)$. We then apply a row-major vectorization function, $\operatorname{vec}_r(\cdot)$, to each gradient tensor and concatenate the resulting vectors using a collector operator, $\operatorname{col}(\cdot)$. This forms the full flattened gradient vector $\mathbf{v}_\theta(s_i) \in \mathbb{R}^{M}$, where $M$ is the total number of parameters:
\begin{equation*}
\mathbf{v}_\theta(s_i) \coloneqq \operatorname{col}\left( \left\{\operatorname{vec}_r(\nabla_{\mathbf{W}^{(l,k)}}\mathcal L(s_i,\theta_T))\right\}_{(l,k) \in \mathcal{W}} \right)
\end{equation*}
\FloatBarrier
\subsection{Cosine Similarity Reconstruction via Dot Products}\label{app:cosine_reconstruction}
The standard dot product $\mathbf{a}^{\top}\mathbf{b}$ is sensitive to both the angle between vectors and their magnitudes. Since gradient magnitudes can vary significantly across samples \citep{DBLP:conf/nips/SuiWS21}, a large norm could dominate the score and mask the true geometric alignment. To create a magnitude-invariant measure, we use cosine similarity, which isolates the directional component of influence:
\begin{equation*}
\text{sim}_{\text{cos}}(\mathbf{a},\mathbf{b}) = \frac{\mathbf{a}^{\top}\mathbf{b}}{\|\mathbf{a}\|_2\,\|\mathbf{b}\|_2}.
\end{equation*}
A key property for our \textbf{Select} method is that the cosine similarity for any subset of components $\mathcal{S} \subseteq \mathcal{W}$ can be perfectly reconstructed by summing pre-computed, component-wise dot products. For a query sample $s_{q_i}$ and a candidate sample $s_j$, we first define the pre-computed scalar dot products for each component $(l,k) \in \mathcal{W}$ as:
\begin{equation*}
\delta^{(l,k)}_{i,j} \coloneqq \mathbf{v}_{\mathbf{W}^{(l,k)}}(s_{q_i})^{\top}\mathbf{v}_{\mathbf{W}^{(l,k)}}(s_j).
\end{equation*}
The cosine similarity for the subset $\mathcal{S}$ is then reconstructed by summing these values:
\begin{equation*}
\begin{split}
\text{sim}_{\text{cos}}(\mathbf{v}_{\mathcal{S}}(s_{q_i}), \mathbf{v}_{\mathcal{S}}(s_j)) = \\ \frac{\sum_{(l,k)\in\mathcal{S}} \delta^{(l,k)}_{i,j}}{\sqrt{\sum_{(l,k)\in\mathcal{S}} \delta^{(l,k)}_{i,i}} \sqrt{\sum_{(l,k)\in\mathcal{S}} \delta^{(l,k)}_{j,j}}}.
\end{split}
\end{equation*}
This allows us to efficiently evaluate any subset of components without storing or re-computing high-dimensional gradient vectors.

\section{Algorithmic Details}\label{app:algo_details}
\FloatBarrier
\subsection{Query Dataset Construction}\label{app:dataset_construction}
\paragraph{Paraphrasing prompt}
To create the \textit{paraphrased} query set ($D_p$), each sample from the LIMA subset $D$ was rewritten using GPT-4o-mini, guided by the system prompt in Figure \ref{system_prompt}.
\begin{figure}[ht]
\centering
\small
\begin{minipage}[t]{0.48\textwidth}
\noindent\texttt{You are a paraphrasing expert who is 
specialized in rewriting text 
(questions, statements, etc.) without 
altering the content.Keep in mind, 
that the meaning must not change 
after the paraphrasing. Just output 
the paraphrased text without any 
additional information.
}
\end{minipage}
\caption{System prompt for $D_p$}\label{system_prompt}
\end{figure}
\FloatBarrier
\onecolumn
\subsection{Candidate Set Creation}
\label{app:candidate_creation}
To make the retrieval task computationally tractable, we pre-filter the search space for each query sample $s_{q_i}$. Instead of comparing its gradient against all $N$ samples in the original dataset $D$, we identify a small candidate set $\mathcal{C}_i(b)$ containing the indices of the top-$b$ most lexically similar samples using the BM25 ranking function \citep{DBLP:journals/ftir/RobertsonZ09}. This significantly reduces the number of expensive gradient computations from $\mathcal{O}(N)$ to $\mathcal{O}(b)$ per query. Algorithm~\ref{alg:bm25_select_samples} details this process. To ensure the correct original counterpart $s_i$ is always included for evaluation, we add its index to the set if it is not already present among the top-$b$ candidates, replacing the least similar one.

\begin{algorithm}[ht!]
\small
\caption{Candidate Set Creation via BM25}
\label{alg:bm25_select_samples}
\KwData{Query sample $s_{q_i}$; Original dataset $D=\{s_j\}_{j=1}^{N}$; Set size $b$}
\KwResult{Set of candidate indices $\mathcal{C}_i(b)$}
\BlankLine
\For{$j \gets 1$ \KwTo $N$}{
    $\text{score}[j] \gets \operatorname{BM25}(s_{q_i}, s_j)$\;
}
$\mathcal{C}_i(b) \gets \operatorname{argsort\_top}(\text{score}, b)$\; \tcp*{Get indices of top $b$ scores}
\BlankLine
\tcp{Ensure original counterpart's index $i$ is present}
\If{$i \notin \mathcal{C}_i(b)$}{
    $j_{min} \gets \arg\min_{j' \in \mathcal{C}_i(b)} \text{score}[j']$\; \tcp*{Find index of least similar candidate}
    $\mathcal{C}_i(b) \gets (\mathcal{C}_i(b) \setminus \{j_{min}\}) \cup \{i\}$\; \tcp*{Replace it with index $i$}
}
\Return $\mathcal{C}_i(b)$
\vspace{0.5cm}
\end{algorithm}

\FloatBarrier

\subsection{Greedy Selection by Accuracy}
\label{app:greedy_algorithm}
The \textit{Select} method employs a greedy forward algorithm to find a subset of components that maximizes retrieval accuracy. Algorithm~\ref{alg:greedy_layer_selection_accuracy} provides a formal specification. The algorithm iteratively adds the component that yields the greatest improvement to the accuracy metric, operating entirely on the pre-computed dot products $\delta^{(l,k)}_{i,j}$. We introduce the notation $D^{\mathcal{S}}_{i,j} \coloneqq \sum_{(l,k)\in\mathcal{S}}\delta^{(l,k)}_{i,j}$ to denote the accumulated dot product for a selected set $\mathcal{S}$.

\begin{algorithm*}[ht!]
\small
\caption{Forward Greedy Selection by Retrieval Accuracy}
\label{alg:greedy_layer_selection_accuracy}
\KwData{
Candidate index sets $\{\mathcal{C}_i(b)\}_{i=1}^{N}$;
Pre-computed dot products $\{\delta^{(l,k)}_{i,j}\}$;
A small constant $\varepsilon>0$.
}
\KwResult{Ordered list of (component, accuracy) pairs.}
\BlankLine
Initialize accumulated dot products $D^{\varnothing}_{i,j}\gets 0$ for all required pairs $(i,j)$.\;
$\mathcal{S}\gets\varnothing$ (selected set), $\mathcal{R}\gets\mathcal{W}$ (remaining set).\;
\For{$t \gets 1$ \KwTo $|\mathcal{W}|$}{
  $a^\star\gets -\infty$, $(l^\star,k^\star)\gets\text{None}$.\;
  \ForEach{$(l,k)\in\mathcal{R}$}{
    Let $\mathcal{S}' = \mathcal{S} \cup \{(l,k)\}$.\;
    Compute reconstructed similarities $\hat{\gamma}^{\mathcal{S'}}_{i,j} \gets \frac{D^{\mathcal{S'}}_{i,j}}{\sqrt{D^{\mathcal{S'}}_{i,i}}\sqrt{D^{\mathcal{S'}}_{j,j}}+\varepsilon}$.\;
    $a \gets \frac{1}{N}\sum_{i=1}^{N}\mathbb{I}\big[\hat{\gamma}^{\mathcal{S'}}_{i,i}>\max_{j\in\mathcal{C}_i(b)\setminus\{i\}}\hat{\gamma}^{\mathcal{S'}}_{i,j}\big]$.\;
    \If{$a > a^\star$}{
      $a^\star\gets a$, $(l^\star,k^\star)\gets(l,k)$.\;
    }
  }
  $\mathcal{S}\gets \mathcal{S}\cup\{(l^\star,k^\star)\}$, $\mathcal{R}\gets \mathcal{R}\setminus\{(l^\star,k^\star)\}$.\;
  Update accumulators $D^{\mathcal{S}}_{i,j} \gets D^{\mathcal{S}\setminus\{(l^\star,k^\star)\}}_{i,j} + \delta^{(l^\star,k^\star)}_{i,j}$.\;
  \textbf{Store} $((l^\star,k^\star), a^\star)$ in the output list.\;
}
\Return{Ordered list.}
\vspace{0.5cm}
\end{algorithm*}
\twocolumn
\FloatBarrier

\end{document}